\documentclass[10pt,twocolumn,letterpaper]{article}
\usepackage[dvipsnames,table]{xcolor}
\usepackage[pagenumbers]{cvpr}      

\usepackage[accsupp]{axessibility}

\usepackage[dvipsnames]{xcolor}

\definecolor{cvprblue}{rgb}{0.21,0.49,0.74}
\usepackage[pagebackref,breaklinks,colorlinks,citecolor=cvprblue]{hyperref}

\def\paperID{1887} 
\def\confName{CVPR}
\def\confYear{2025}

\title{\model{}: Decentralized Federated Learning via Mixture of Experts for Medical Data Analysis}

\newcommand{\authorskip}{\hspace{12mm}}
\author{
 Luyuan Xie$^{1,2,3}$\thanks{This work was supported by the National Key R$\&$D Program of China under Grant No.2022YFB2703301.} \authorskip Tianyu Luan$^{4}$\thanks{Tianyu Luan is corresponding author: tianyulu@buffalo.edu} \authorskip Wenyuan Cai$^{1}$ \authorskip
 Guochen Yan$^{2,3}$ \authorskip
 Zhaoyu Chen$^{1,2,3}$ \\ Nan Xi$^{4}$ \authorskip Yuejian Fang$^{1,2,3}$ \authorskip Qingni Shen$^{1,2,3}$ \authorskip Zhonghai Wu$^{1,2,3}$ \authorskip Junsong Yuan$^{4}$\\[3mm]
 {\small $^1$School of Software and Microelectronics, Peking University ~~~~~~ $^2$PKU-OCTA Laboratory for Blockchain and Privacy Computing} \\
  {\small $^3$ National Engineering Research Center for Software Engineering, Peking University~~~~~~ $^4$State University of New York at Buffalo} \\
}

\usepackage{times}
\usepackage{epsfig}
\usepackage{graphicx}
\usepackage{amsmath}
\usepackage{amssymb}
\usepackage{enumitem}

\usepackage{booktabs}
\usepackage{color}
\usepackage{diagbox}
\usepackage{microtype}
\usepackage{makecell}

\usepackage{colortbl}
\usepackage{url}

\usepackage{amsmath,amsfonts}
\usepackage{algorithmic}
\usepackage{graphicx}
\usepackage{textcomp}
\usepackage{xspace}
\usepackage{xcolor}
\usepackage{threeparttable}
\usepackage{booktabs}
\usepackage{url}
\usepackage{multirow} 
\usepackage{pifont}
\usepackage{algorithm}
\usepackage{algorithmic}
\usepackage{makecell}
\usepackage{newfloat}
\usepackage{listings}
\usepackage{etoolbox}
\usepackage{latexsym}
\usepackage{times}
\usepackage{xcolor}
\usepackage{colortbl}
\usepackage{newfloat}
\usepackage{listings}
\usepackage{newfloat}
\usepackage{listings}

\usepackage{amssymb}
\usepackage{pifont}
\usepackage{graphicx}
\usepackage{latexsym}
\usepackage{times}
\usepackage{soul}
\usepackage{url}
\usepackage{booktabs}
\usepackage{algorithm}
\usepackage{algorithmic}
\usepackage{multirow}
\usepackage{enumitem}

\usepackage{amsmath}
\usepackage{amssymb}
\usepackage{mathtools}
\usepackage{amsthm}

\usepackage[capitalize,noabbrev]{cleveref}
\usepackage{microtype}
\usepackage{graphicx}
\usepackage{booktabs} 
\usepackage{booktabs}
\usepackage{cleveref}

\def\model{SAUCD}

\usepackage[capitalize]{cleveref}
\crefname{section}{Sec.}{Secs.}
\Crefname{section}{Section}{Sections}
\Crefname{table}{Table}{Tables}
\crefname{table}{Tab.}{Tabs.}

\makeatletter

\newcommand{\Rmnum}[1]{\textcolor{red}{\expandafter\@slowromancap\romannumeral #1@}}

\makeatother

\def\model{dFLMoE}

\begin{document}
\maketitle
\begin{abstract}

Federated learning has wide applications in the medical field. It enables knowledge sharing 
among different healthcare institutes while protecting patients' privacy. However, existing federated learning systems are typically centralized, requiring clients to upload client-specific knowledge to a central server for aggregation. This centralized approach would integrate the knowledge from each client into a centralized server, and the knowledge would be already undermined during the centralized integration before it reaches back to each client. Besides, the centralized approach also creates a dependency on the central server, which may affect training stability if the server malfunctions or connections are unstable. To address these issues, we propose a decentralized federated learning framework named \model{}. In our framework, clients directly exchange lightweight head models with each other. After exchanging, each client treats both local and received head models as individual experts, and utilizes a client-specific Mixture of Experts (MoE) approach to make collective decisions. This design not only reduces the knowledge damage with client-specific aggregations but also removes the dependency on the central server to enhance the robustness of the framework. We validate our framework on multiple medical tasks, demonstrating that our method evidently outperforms state-of-the-art approaches under both model homogeneity and heterogeneity settings.

\end{abstract}


\section{Introduction}
\label{sec:intro}

Federated learning has extensive medical applications. A well-designed federated system can protect data privacy while sharing high-level knowledge among different clients. This enables each client's network to receive additional support and achieve better performance and generalizability. In medical scenarios, patient data is hard to collect and has strong privacy protection requirements. Federated learning systems can effectively address the data limitations at each healthcare institution, enhancing their model performance and generalizability while ensuring privacy.

\begin{figure}[t]
  \centering
   \includegraphics[width=1\linewidth]{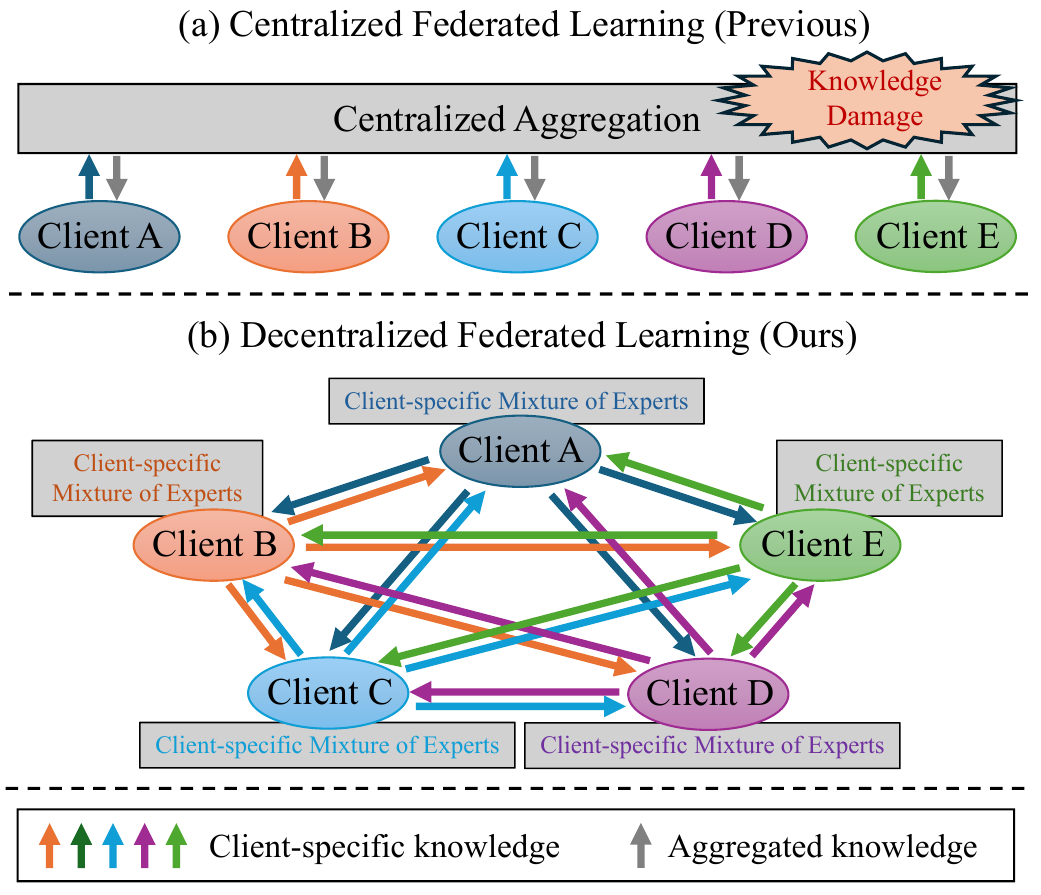}
   \caption{(a) Previous centralized federated learning framework aggregates knowledge from each client in a centralized server. This process can lead to knowledge damage in centralized aggregation and the framework is heavily dependent on the central server's stability. (a) Our decentralized framework \model{} eliminates centralized server and aggregation by having clients directly exchange knowledge with each other. Each client then uses a Mixture of Experts (MoE) approach to adaptively combine local and received knowledge.}
   \label{fig:teaser}
\end{figure}

Existing federated learning systems, such as \cite{fedavg,fedrep,apfl,lg-fedavg}, are designed in a centralized manner. In each training round, each client needs to upload client-specific knowledge (\eg{} model parameters) to a central server for aggregation, which is then distributed back to each client. Regarding aggregation methods, they require a unified model structure \cite{fedavg,fedrep,apfl}, a centralized messenger model \cite{xie2024mh,DBLP:journals/corr/abs-2311-06879,xie2024mhpfl}, or a unified public dataset \cite{DBLP:journals/corr/abs-1910-03581,9392310,NEURIPS2020_18df51b9,NEURIPS2021_5383c731}. Such centralized designs achieve good results, but this design may lead to performance bottlenecks. As shown in \cref{fig:teaser}(a), centralized federated learning frameworks, such as \cite{NEURIPS2021_5383c731,fedavg}, would distill knowledge from each client from their local data, and then send that knowledge to a centralized server which aggregates that knowledge into a single model. However, the aggregation process would typically merge the information from all client models into a single aggregated model, mostly with a sample merging scheme such as weighted sum. Considering the domain and data distribution differences among clients, the same aggregation process for all clients would result in potential knowledge damage even before the aggregated knowledge gets back to each client. Moreover, such centralized aggregation methods, particularly weighted sum schemes, are also widely used in federated systems like \cite{fedavg,karimireddy2020scaffold,fedprox}, which may not preserve the knowledge of each client well and could possibly hurt the performance of the federated learning framework. Furthermore, centralized federated frameworks heavily depend on the central server and the stability of its connections. If the central server malfunctions or the connections to it are unstable, the training stability of each client can be significantly affected.

To address the knowledge damage of centralized aggregation and to reduce the dependence on the centralized server, we propose a decentralized approach to design a federated learning framework. As illustrated in \cref{fig:teaser}(b), to minimize knowledge damage during model aggregation, we eliminate the centralized model aggregation operation. Instead, during the knowledge exchange process, the knowledge that each client would originally send to the server is now directly transmitted to other clients. This way, each client can receive the full knowledge sent by others without damage. Note that the communications between clients do not involve any patient data, which allows us to effectively protect patient privacy. Then, within each client, we design a Mixture of Experts (MoE) approach, treating the knowledge received from other clients and the client's own local knowledge as individual experts, and making decisions collectively using these experts. This decentralized design enables each client to consider its own local data and adaptively select the participation and weights of the experts and also avoids the unnecessary knowledge damage that occurs in centralized systems when aggregating into a unified model. Furthermore, it eliminates the reliance on a central server. If a client or some connections are unstable, the training of our framework would still be effective and without interruption.

Our decentralized system is named Decentralized Federated Learning via Mixture of Experts (\model{}). In each training round, we first train the local model of each client, which consists of a body and a head. Each client's body processes the input and encodes it into features, which are then passed through the head to obtain the final results. After local training, we send the model heads from each client to all other clients. Considering that a decentralized framework requires model transmission between every pair of clients, transmitting only the lightweight head models would significantly reduce communication costs. After obtaining the heads from other clients, we train an attention-based MoE model, adaptively selecting the most effective combination of heads on each client to obtain the final results. Such client-specific MoE design does not require a structure consistency of the head from each client, which can effectively accommodate the commonly occurring model heterogeneity in practical medical scenarios. Moreover, due to the decentralized nature of the system, when a certain client encounters issues, other clients can still be trained without interruption. If the connection between two clients drops, the knowledge from these clients can still be shared through others, enhancing the robustness of the framework.

In summary, our contributions are as follows:

\begin{itemize}[noitemsep,topsep=0pt]
    \item We propose a decentralized federated learning framework named \model{}. Our framework directly transmits each client's knowledge to other clients and performs local decision-making on each client, effectively avoiding the knowledge damage caused by centralized server aggregation and eliminating the dependence on a central server.
    \item We design a lightweight Mixture of Experts (MoE) module for each client. This local MoE module can adaptively make client-specific decisions using lightweight experts from local and other clients, which can better adapt knowledge from other clients to improve performance and generalizability, without notably increasing communication costs.
    \item We validate the effectiveness of our framework on 5 different medical tasks. Our experimental results demonstrate that, on these tasks, under both model homogeneity and heterogeneity settings, our method evidently outperforms the state-of-the-art.
\end{itemize}

\section{Related Works}
\label{sec:related}
\textbf{Centralized federated learning.}
The general paradigm of federated learning involves clients uploading their local knowledge to a central server for aggregation, which is then distributed back to all clients. Based on the type of aggregation methods, this can be divided into three main categories: local model parameters aggregation \cite{fedavg,karimireddy2020scaffold,fedprox,clustered1,clustered2,li2021ditto,apfl, feng2023privacy, yan2024fedvck, yan2024toward}, soft predictions aggregation \cite{DBLP:journals/corr/abs-1910-03581,9392310,NEURIPS2020_18df51b9,NEURIPS2021_5383c731,gong2021ensemble,gong2022preserving,gong2022federated,gong2024federated,chen2024federated}, and messenger model parameters aggregation \cite{xie2024mh, xie2024mhpfl, DBLP:journals/corr/abs-2311-06879, DBLP:journals/corr/abs-2310-13283}. 
The framework for local model parameters aggregation requires aggregating all or part of the local model parameters at the central server \cite{lcfed,li2021ditto,mtl1,mtl2,mtl3,pmlr-v162-marfoq22a,fedrep,lg-fedavg}. They require consistent local model structure \cite{fedper,fedhealth,knnper,xie2024pflfe}. Federated learning frameworks that aggregate soft predictions require a public dataset, limiting their application in medical scenarios.  Frameworks based on aggregating messenger model parameters insert a homogeneous model into each client and share this model to transfer knowledge. These centralized approaches can lead to knowledge damage during aggregation, and the knowledge would be undermined before it reaches back to each client. Meanwhile, if the central server malfunctions or the connections are unstable, the training stability of each client can be significantly impacted.


\textbf{Decentralized federated learning.}
Decentralized federated learning, also known as peer-to-peer federated learning \cite{roy2019braintorrent}, addresses the dependency on a central server. Currently, mainstream research in decentralized federated learning focuses on integrating it with blockchain to further enhance security and privacy \cite{liu2020fedcoin,agarwal2024towards,qin2024blockdfl,zhao2019mobile}. However, these works did not address the statistical heterogeneity and system heterogeneity issues in federated learning. Meanwhile, recent work \cite{li2022towards,chen2022cfl,wang2021non,salmeron2023benchmarking} has emerged to improve the performance of decentralized federated learning. 
They have only decentralized the security aspect without decentralizing the algorithm. Our method adopts a localized knowledge fusion approach, allowing us to adaptively select knowledge based on each client's needs, thereby reducing knowledge damage.


\section{Method}
\label{sec:method}
\subsection{Overview}
We design a decentralized federated learning framework named \model{} to address the knowledge damage of centralized aggregation and reduce the dependence on the centralized server.
Specifically, we firstly train a local network for client $i$ by its private dataset $D_i = \{x_{i}, y_{i}\}$, where $x_{i}$ is the input data in $D_i$, and $y_{i}$ is the label. Then each client shares their learned knowledge $K$ with other clients. Finally, we achieve the final decision through knowledge fusion using Mixture of Experts (MoE). The \model{}’s paradigm can be expressed as: 
\begin{equation} 
\mathbb{G} = \bigcup^N_{i=0}{f_i(\theta_i ; x_i; \{K_1,\cdots K_i,\cdots K_N\})},
\end{equation} 
where $f_i(\theta _i; x_i; \{K_1,\cdots K_i,\cdots K_N\}) $ is the model for client $i$, where $\theta _i$ is the parameters of $f_i$, $x_i$ is model input, $K_i$ is the knowledge shared by client $i$, $N$ represents the total number of participating clients, and $\mathbb{G}$ represents the the set of $f_i$. 



\begin{figure}[t]
\includegraphics[width=1\linewidth]{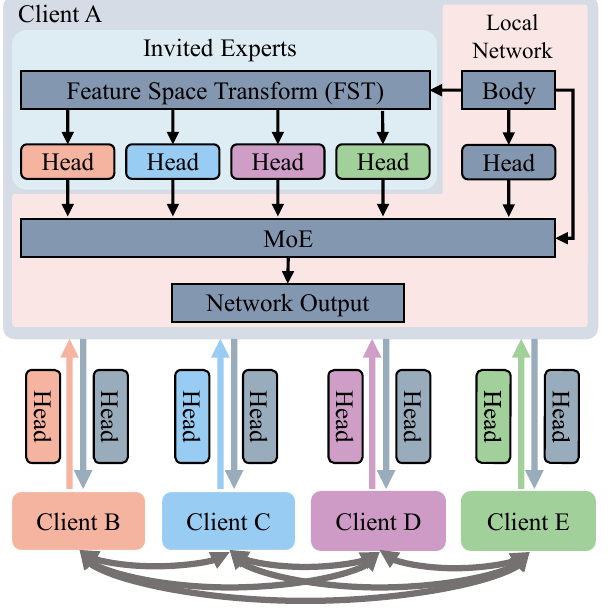}
\centering
\caption{Overview of our proposed \model{} framework. For Each training phase, we first train the Local network (Body and Head) while freezing the parameters of the MoE module (top right). Then, we send and receive the head to share knowledge among clients (bottom). Finally, we do a Mixture-of-Experts (MoE) decision by training the Feature space transform and MoE network while freezing other parameters including the local body and all the heads. More details can be found in the \cref{sec:method}.}
\label{fig:pipeline}
\end{figure}


The pipeline of \model{} is shown in \cref{fig:pipeline}. 
In each client, the model includes the local network and invited experts. The local network is divided into four parts: Body, Feature space transform, Head, and Mixture of Experts (MoE). The body model is used to extract features. Feature space transform module converts the local features into the feature space corresponding to the respective experts (heads). The head model generates the network output using the features and the head module of each client is also shared among all clients, with the heads invited from other clients forming the Mixture of Experts module for each client. We treat each head as an expert and use a Mixture of Experts (MoE) approach to get the final outputs. Our training process consists of 3 steps: a) Local network training, b) Sharing the local head among clients, and c) Mixture of Experts decision. In the rest of the section, we will explain each step in detail.



\subsection{Local Network Training} 
At this stage, our goal is to obtain the local network with local knowledge by local data. Therefore, we only train the head and body of the local network and freeze the parameters of the feature space transform and MoE module. For the client $i$, the local network output $\hat{y}^l_{i}$ can be defined as:
\begin{equation} 
\hat{y}^l_{i} = F_{h,i}(F_{b,i}(x_{i})),
\label{eq:yl1}
\end{equation}
where $F_{b,i}(\cdot)$ and $F_{h,i}(\cdot)$ are the body and head of the local network in client $i$, respectively.  The MoE output $\hat{y}^m_{i}$ can be represented as:
\begin{equation} 
\hat{y}^m_{i} = M(\bigcup^N_{j=1}F_{h,j}(FT_j(F_{b,i}(x_{i}))), F_{b,i}(x_{i}))),
\label{eq:ym1}
\end{equation}
where $M(\cdot)$ is the MoE network and $FT_j$  is the feature space transform of Experts $j$ (See \cref{sec:irit} for details) in Fig.2, $F_{b,i}(\cdot)$ is the local model body, $\bigcup^N_{j=1}F_{h,j}(FT_j(F_{b,i}(x_{i})))$ is the set of
 predictions from each expert (head). $N$ is the total number of participating clients, excluding the local client.

Finally, for client $i$, its training loss function $\mathcal{L}_{\textit{ln},i}$ is:
\begin{equation} 
{\mathcal{L}_{\textit{ln},i}} = \lambda_{\textit{loc}}{\mathcal{L}_{\textit{loc}}(\hat{y}^l_{i},y_{i})}  + {\lambda_{\textit{MoE}}} {\mathcal{L}_{\textit{MoE}}(\hat{y}^m_{i},y_{i})}.
\label{eq:l_inji}
\end{equation} 
 $\mathcal{L}_{\textit{loc}}$ and $\mathcal{L}_{\textit{MoE}}$  represent the loss functions of the local network and the MoE, respectively. For classification tasks, $\mathcal{L}_{\textit{loc}}$ and $\mathcal{L}_{\textit{MoE}}$ are cross-entropy loss. For super-resolution tasks, $\mathcal{L}_{\textit{loc}}$ and $\mathcal{L}_{\textit{MoE}}$ are $L1$ loss. And for segmentation tasks, $\mathcal{L}_{\textit{loc}}$ and $\mathcal{L}_{\textit{MoE}}$ are Dice and cross-entropy loss. $\lambda_{\textit{loc}}$ and $\lambda_{\textit{MoE}}$ are their corresponding weights. $y_{i}$ is the label of local data $x_{i}$. More details can be found in supplementary materials.


\subsection{Localized knowledge Exchange} 
In the communication stage of existing decentralized federated learning, each of the $N$ participating clients needs to share its local model with the other $N-1$ clients. Thus, in total there are $N(N-1)$ times of communication, which is significantly higher than the centralized federated learning, which would only need $2N$ communications for both uploading and downloading. To reduce the communication cost of decentralized federated learning, in the Sharing local head among clients phase, we only share the head of the local model instead of the entire local model. The parameters of the head are several orders of magnitude smaller than those of the local model, which significantly reduces computational costs. Compared to centralized federated learning, this approach does not introduce a significant communication burden. Our experiments demonstrate that, in contrast to sharing the entire local model with each client, our communication overhead is only 0.02$\%$ of theirs, while our performance remains comparable to theirs.



\begin{figure}[t]
\includegraphics[width=1\linewidth]{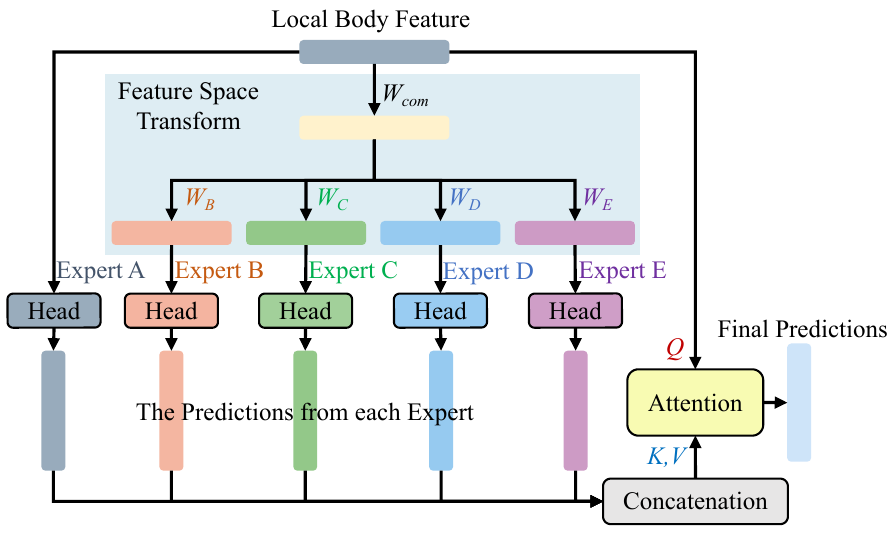}
\centering
\caption{The structure of  Mixture of Experts and Feature Space Transform. Firstly, the Feature Space Transform converts the local body feature into the feature space corresponding to each expert. Then, each feature obtains the final prediction through the respective expert, and we collect all predictions as the Key $K$ and Value $V$.
Next, we generate the query $Q$ using the local body feature through a linear layer. Finally, we perform the attention mechanism with $Q$, $K$, and $V$ to obtain the final predictions.}
\label{fig_3}
\end{figure}


\subsection{Mixture of Experts Decision} 
This stage is designed to learn the combination weights of all experts based on the local data. During this stage of training, we fine-tune the parameters of feature space transform and MoE while freezing other parameters. The  Mixture of Experts Decision loss function  $\mathcal{L}_{\textit{MD},i}$ for client $i$ is defined as:
\begin{equation} 
{\mathcal{L}_{\textit{MD},i}} =\mathcal{L}_{\textit{MoE}}(\hat{y}^m_{i},y_{i}),
\label{eq:l_diff}
\end{equation} 
where $\mathcal{L}_{\textit{MoE}}$ is the loss functions of the MoE. For classification tasks, $\mathcal{L}_{\textit{MoE}}$ is cross-entropy loss. For super-resolution tasks, $\mathcal{L}_{\textit{MoE}}$ is $L1$ loss. And for segmentation tasks, $\mathcal{L}_{\textit{MoE}}$ is Dice and cross-entropy loss. During inference, we directly use the output of the MoE as the final prediction. The experimental results show that \model{} can be applied to federated learning scenarios with data heterogeneity, model homogeneity, and model heterogeneity without notably increasing communication costs.




\label{sec:irit}

\begin{table*}[htbp]
  \centering
  \caption{The results of classification task in different resolutions with homogeneous models or heterogeneous models. The x2↓, x4↓, and x8↓ are downsampling half, quarter, and eighth of high-resolution images. We evaluate ACC and MF1 results on the BreaKHis dataset. The larger the better. \textbf{Bold} number means the best. The {\color{red!70}red} boxes represent the single model federated learning and personalized federated learning methods, and their individual clients use the homogeneous model settings (ResNet5). The {\color{blue!70}blue} boxes represent the method of using heterogeneous models. The four client models are set to ResNet$\lbrace 17,11,8,5 \rbrace$, respectively. In two different model settings, \model{} achieves the best performance.}
  \resizebox{1\linewidth}{!}{
    \begin{tabular}{c|cc|cc|cc|cc|cc}
    \hline
    \multirow{2}{*}{Methods} & \multicolumn{2}{c|}{HR} & \multicolumn{2}{c|}{x2} & \multicolumn{2}{c|}{x4} & \multicolumn{2}{c|}{x8} & \multicolumn{2}{c}{Average} \\
\cline{2-11}          & ACC$\uparrow$   & MF1$\uparrow$   & ACC$\uparrow$   & MF1$\uparrow$   & ACC$\uparrow$   & MF1$\uparrow$   & ACC$\uparrow$   & MF1$\uparrow$   & ACC$\uparrow$   & MF1$\uparrow$ \\
    \hline
   \rowcolor{red!10} Only Local Training & 0.7491  & 0.6719  & 0.7568 & 0.6856 & 0.7015 & 0.6135 & 0.6956  & 0.5867  & 0.7258  & 0.6394  \\
    \hline
   \rowcolor{red!10} FedAvg & 0.6067  & 0.4621  & 0.6667  & 0.5874  & 0.6178  & 0.5194  & 0.5799  & 0.4616  & 0.6178  & 0.5076  \\
   \rowcolor{red!10} SCAFFOLD & 0.6263  & 0.4821  & 0.7156  & 0.6597  & 0.6475  & 0.5906  & 0.5702  & 0.4969  & 0.6399  & 0.5573  \\
   \rowcolor{red!10} FedProx & 0.6195  & 0.4958  & 0.6862  & 0.6271  & 0.6467  & 0.5632  & 0.4664  & 0.3495  & 0.6047  & 0.5089  \\
    \hline
   \rowcolor{red!10} Ditto & 0.7111  & 0.6557  & 0.7321  & 0.6404  & 0.7261  & 0.6743  & 0.6854  & 0.5932  & 0.7137  & 0.6409  \\
   \rowcolor{red!10} APFL  & 0.6412  & 0.5848  & 0.6033  & 0.5626  & 0.7301  & 0.6468  & 0.6973  & 0.6166  & 0.6680  & 0.6027  \\
   \rowcolor{red!10} FedRep & 0.7663  & 0.7165  & 0.7513  & 0.6869  & 0.6849  & 0.6151  & 0.7254  & 0.6229  & 0.7320  & 0.6604  \\
   \rowcolor{red!10} LG-FedAvg & 0.7358  & 0.6504  & 0.7733  & 0.6726  & 0.7182  & 0.6323  & 0.7173  & 0.6481  & 0.7362  & 0.6509  \\
    \hline
   \rowcolor{red!10} MH-pFLID & 0.8282  & 0.7762  & 0.8308  & 0.7829  & 0.8180  & 0.7674  & 0.7560  & 0.6933  & 0.8083  & 0.7550  \\
    \hline
   \rowcolor{red!10} \model{} (Ours) & \textbf{0.8652} & \textbf{0.8360} & \textbf{0.8597} & \textbf{0.8322} & \textbf{0.8423} & \textbf{0.8063} & \textbf{0.7602} & \textbf{0.7131} & \textbf{0.8319} & \textbf{0.7969} \\
    \hline
   \rowcolor{blue!10} Only Local Training & 0.7891  & 0.7319  & 0.8027 & 0.7461 & 0.7538 & 0.6852 & 0.6956  & 0.5867  & 0.7603  & 0.6875  \\
    \hline
   \rowcolor{blue!10} FedMD & 0.7599  & 0.7083  & 0.8321  & 0.7829  & 0.7721  & 0.7293  & 0.6495  & 0.5439  & 0.7534  & 0.6911  \\
   \rowcolor{blue!10} FedDF & 0.7661  & 0.7253  & 0.8132  & 0.7629  & 0.7826  & 0.7342  & 0.6627  & 0.5627  & 0.7562  & 0.6963  \\
   \rowcolor{blue!10} pFedDF & 0.8233  & 0.7941  & 0.8369  & 0.7965  & 0.8121  & 0.7534  & 0.6843  & 0.6022  & 0.7892  & 0.7366  \\
   \rowcolor{blue!10} DS-pFL & 0.7842  & 0.7609  & 0.8334  & 0.7967  & 0.7782  & 0.7258  & 0.6327  & 0.5229  & 0.7571  & 0.7016  \\
   \rowcolor{blue!10} KT-pFL & 0.8424  & 0.8133  & 0.8441  & 0.8011  & 0.7801  & 0.7325  & 0.7032  & 0.6219  & 0.7925  & 0.7422  \\
  \hline
   \rowcolor{blue!10} MH-pFLID & 0.8929  & 0.8658  & 0.8992  & 0.8787  & 0.8661  & 0.8327  & 0.7751  & 0.7130  & 0.8583  & 0.8226  \\
    \hline
   \rowcolor{blue!10} \model{} (Ours) & \textbf{0.9048 } & \textbf{0.8898} & \textbf{0.9205} & \textbf{0.9064} & \textbf{0.9039} & \textbf{0.8865} & \textbf{0.8227} & \textbf{0.7819} & \textbf{0.8880} & \textbf{0.8662 } \\
    \hline
    \end{tabular}%
  \label{tab:drc}%
  }
\end{table*}%

\begin{table}[htbp]
  \centering
  \caption{The results of super-resolution with homogeneous models or heterogeneous models. The x8$\uparrow$, x4$\uparrow$, and x2$\uparrow$ are two times, four times, and eight times super-resolution for downsampling eighth, quarter, and half of high-resolution images. We evaluate PSNR and SSIM results on the BreaKHis dataset. The larger the better.  The {\color{red!70}red} boxes represent the method of individual clients adopting the homogeneous model settings (RCNN). The {\color{blue!70}blue} boxes represent the method of using heterogeneous models. The three client models are set to SRResNet$\lbrace 18, 12, 6 \rbrace$, respectively. In two different model settings, \model{} achieves the best performance.}
   \resizebox{1\linewidth}{!}{
\begin{tabular}{c|cc|cc|cc|cc}
    \hline
    \multirow{2}{*}{Method} & \multicolumn{2}{c|}{x8$\uparrow$} & \multicolumn{2}{c|}{x4$\uparrow$} & \multicolumn{2}{c|}{x2$\uparrow$} & \multicolumn{2}{c}{Average} \\
\cline{2-9}          & PSNR$\uparrow$  & SSIM$\uparrow$  & PSNR$\uparrow$  & SSIM$\uparrow$  & PSNR$\uparrow$  & SSIM$\uparrow$  & PSNR$\uparrow$  & SSIM$\uparrow$ \\
    \hline
    \rowcolor{red!10} Bicubic & 20.75  & 0.4394  & 23.21  & 0.6305  & 26.60  & 0.9151  & 23.52  & 0.6617  \\
    \hline
    \rowcolor{red!10} Only Local Training & 21.12  & 0.4872  & 24.04  & 0.6634  & 28.36  & 0.8631  & 24.51  & 0.6712  \\
    \hline
   \rowcolor{red!10} FedAvg & 22.00  & 0.6572  & 24.65  & 0.6802  & 26.46  & 0.8188  & 24.37  & 0.7187  \\
   \rowcolor{red!10} SCAFFCOLD & 21.33  & 0.5633  & 24.47  & 0.6817  & 28.61  & 0.8398  & 24.80  & 0.6949  \\
   \rowcolor{red!10} FedProx & 21.77  & 0.6254  & 23.92  & 0.6791  & 27.60  & 0.8274  & 24.43  & 0.7106  \\
    \hline
   \rowcolor{red!10} LG-FedAvg & 21.50  & 0.4461  & 23.63  & 0.6789  & 27.02  & 0.8352  & 24.05  & 0.6534  \\
   \rowcolor{red!10} FedRep & 22.01  & 0.6170  & 24.73  & 0.6999  & 29.72  & 0.8964  & 25.49  & 0.7378  \\
    \hline
   \rowcolor{red!10} Ours  & \textbf{23.43} & \textbf{0.6671} & \textbf{27.59} & \textbf{0.8272} & \textbf{34.82} & \textbf{0.9605} & \textbf{28.61}  & \textbf{0.8183}  \\
    \hline
   \rowcolor{blue!10} Only Local Training & 21.76  & 0.5141  & 25.23  & 0.7423  & 29.31  & 0.9022  & 25.43  & 0.7195  \\
    \hline
   \rowcolor{blue!10} Ours  & \textbf{23.94} & \textbf{0.6929} & \textbf{28.08} & \textbf{0.8436} & \textbf{35.87} & \textbf{0.9686} & \textbf{29.30} & \textbf{0.8350} \\
    \hline
    \end{tabular}%

  \label{tab:SR}%
  }
\end{table}%

\begin{figure*}[ht]
\includegraphics[width=1\linewidth]{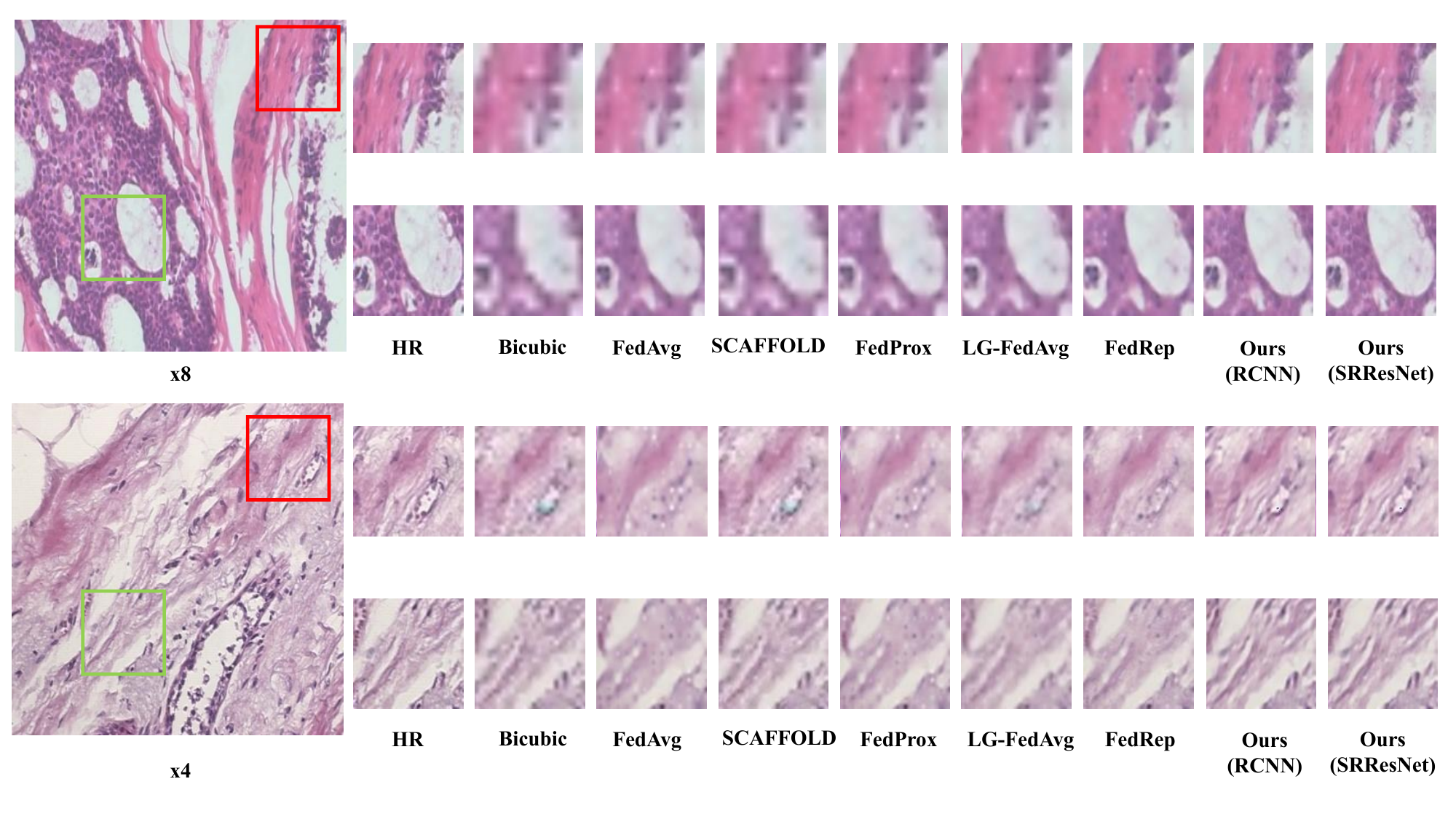}
\centering
\caption{Visualized comparison of Federated Learning in medical image super-resolution. We randomly select two samples from different resolutions (x8↓ and x4↓) to form the visualization. Super-resolution results for FedAVG, SCAFFOLD, FedProx, LG-FedAvg, FedRep,  our method \model{} (RCNN) and \model{} (SRResNet). Our framework can recover more details.}
\label{fig_SR}
\end{figure*}

\begin{figure*}[ht]
\includegraphics[width=0.9\linewidth]{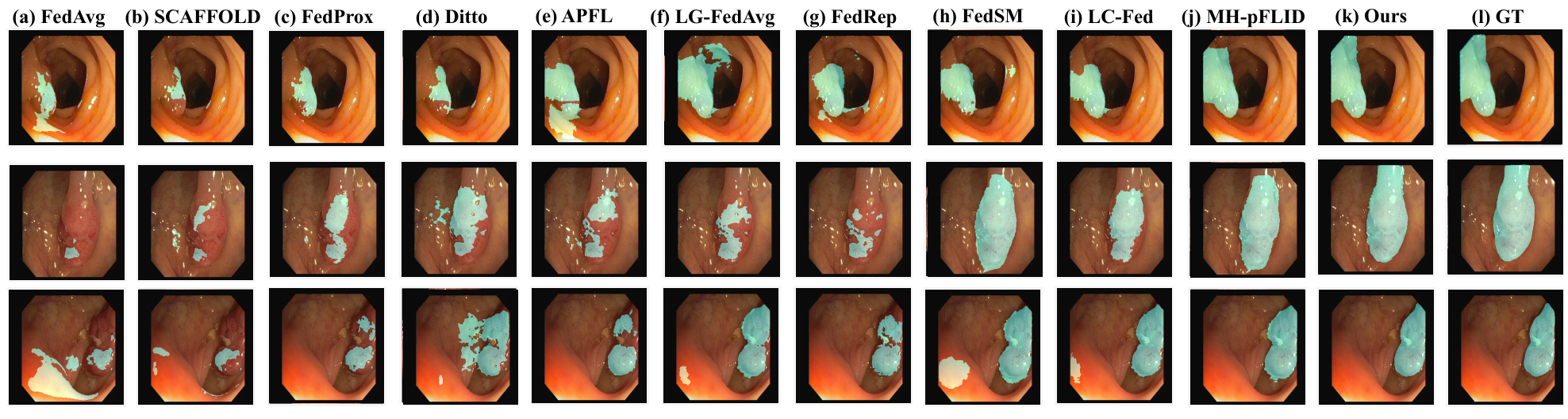}
\centering
\caption{Visualized comparison of Federated Learning in medical image segmentation. We randomly select three samples from different clients to form the visualization. (a-k)  Segmentation results for FedAVG, SCAFFOLD, FedProx, Ditto, APFL, LG-FedAvg, FedRep, FedSM, LC-Fed, MH-FLID and our method \model{}; (l) Ground truths (denoted as ‘GT’).}
\label{fig_seg}
\end{figure*}

\begin{table}[htbp]
  \centering
  \caption{The results of time-series classification with homogeneous models or heterogeneous models.  We evaluate ACC and MF1 results on the Sleep-EDF dataset.  The {\color{red!70}red} boxes represent the method of individual clients adopting the homogeneous model settings (TCN). The {\color{blue!70}blue} boxes represent the method of using heterogeneous models. The three client models are TCN, Transformer, and RNN, respectively. In two different model settings, \model{} achieves the best performance.}
  \resizebox{1\linewidth}{!}{
    \begin{tabular}{c|cc|cc|cc|cc}
     \hline
    \multirow{2}{*}{Method} & \multicolumn{2}{c|}{Client 1} & \multicolumn{2}{c|}{Client 2} & \multicolumn{2}{c|}{Client 3} & \multicolumn{2}{c}{Average} \\
\cline{2-9}          & ACC$\uparrow$   & MF1$\uparrow$   & ACC$\uparrow$   & MF1$\uparrow$   & ACC$\uparrow$   & MF1$\uparrow$   & ACC$\uparrow$   & MF1$\uparrow$ \\
    \hline
   \rowcolor{red!10} Only Local Training & 0.9073 & 0.8757 & 0.8012 & 0.7933 & 0.7791 & 0.7289 & 0.8292  & 0.7993  \\
    \hline
   \rowcolor{red!10} FedAvg & 0.8357 & 0.7281 & 0.7719 & 0.7726 & 0.7418 & 0.6083 & 0.7831  & 0.7030 \\
   \rowcolor{red!10} SCAFFOLD & 0.8792 & 0.8176 & 0.8473 & 0.8494 & 0.7575 & 0.6242 & 0.8280  & 0.7637  \\
   \rowcolor{red!10} FedProx & 0.8541 & 0.7668 & 0.8154 & 0.8162 & 0.7804 & 0.7179 & 0.8166  & 0.7670  \\
    \hline
   \rowcolor{red!10} FedRep &0.8934 & 0.8633 & 0.8367 & 0.8221 & 0.7782 & 0.7341 & 0.8361  & 0.8065  \\
   \rowcolor{red!10} LG-FedAvg &0.8797 & 0.7613 & 0.8532 & 0.8568 & 0.7656 & 0.6954 & 0.8328  & 0.7712  \\
    \hline
   \rowcolor{red!10} MH-pFLID & 0.9392 & 0.9117 & 0.8463 & 0.8321 & 0.8244 & 0.7973 & 0.8700  & 0.8470  \\
    \hline
   \rowcolor{red!10} \model{}(Ours) & \textbf{0.9470} & \textbf{0.9303} & \textbf{0.9201} & \textbf{0.9210} & \textbf{0.8451} & \textbf{0.8123} & \textbf{0.9041} & \textbf{0.8879} \\
    \hline
   \rowcolor{blue!10} Only Local Training & 0.9073 & 0.8757 & 0.8053 & 0.8001 & 0.8012 & 0.7263 & 0.8379  & 0.8007  \\
    \hline
   \rowcolor{blue!10} FedMD & 0.9334 & 0.9225 & 0.7934 & 0.7966 & 0.793 & 0.7072 & 0.8399  & 0.8088  \\
   \rowcolor{blue!10} FedDF & 0.9146 & 0.8893 & 0.7988 & 0.8042 & 0.7881 & 0.6855 & 0.8338  & 0.7930  \\
   \rowcolor{blue!10} pFedDF & 0.9173 & 0.8957 & 0.827 & 0.8309 & 0.8137 & 0.7713 & 0.8527  & 0.8326  \\
   \rowcolor{blue!10} DS-pFL & 0.9133 & 0.9033 & 0.8253 & 0.8301 & 0.8042 & 0.7539 & 0.8476  & 0.8291  \\
   \rowcolor{blue!10} KT-pFL & 0.924 & 0.9089 & 0.8419 & 0.8466 & 0.8204 & 0.7722 & 0.8621  & 0.8426  \\
    \hline
   \rowcolor{blue!10} MH-pFLID & 0.9439 & 0.9248 & 0.8725 & 0.876 & 0.824 & 0.7773 & 0.8801  & 0.8594  \\
    \hline
   \rowcolor{blue!10} \model{}(Ours)  & \textbf{0.9484} & \textbf{0.9319} & \textbf{0.9308} & \textbf{0.9319} & \textbf{0.8617} & \textbf{0.8319} & \textbf{0.9136 } & \textbf{0.8986 } \\
    \hline
    \end{tabular}%
  \label{tab:tsc}%

  }
\end{table}%

\begin{table*}[htbp]
  \centering
  \caption{The results of Image Classification Task with Different Label Distributions. This task includes breast cancer classification and Ocular disease recognition. We evaluate ACC and MF1 results in this task. The larger the better. \textbf{Bold} number means the best. \model{} has the best performance.}
   \resizebox{1\linewidth}{!}{
    \begin{tabular}{c|cc|cc|cc|cc|cc|cc|cc|cc|cc}
    \hline
    \multicolumn{19}{c}{Breast Cancer Classification} \\
    \hline
    \multirow{2}{*}{Method} & \multicolumn{2}{c|}{ResNet} & \multicolumn{2}{c|}{shufflenetv2} & \multicolumn{2}{c|}{ResNeXt} & \multicolumn{2}{c|}{squeezeNet} & \multicolumn{2}{c|}{SENet} & \multicolumn{2}{c|}{MobileNet} & \multicolumn{2}{c|}{DenseNet} & \multicolumn{2}{c|}{VGG} & \multicolumn{2}{c}{Average} \\
\cline{2-19}          & ACC$\uparrow$   & MF1$\uparrow$   & ACC$\uparrow$   & MF1$\uparrow$   & ACC$\uparrow$   & MF1$\uparrow$   & ACC$\uparrow$   & MF1$\uparrow$   & ACC$\uparrow$   & MF1$\uparrow$   & ACC$\uparrow$   & MF1$\uparrow$   & ACC$\uparrow$   & MF1$\uparrow$   & ACC$\uparrow$   & MF1$\uparrow$   & ACC$\uparrow$   & MF1$\uparrow$ \\
    \hline
    Only Local Training & 0.59  & 0.455 & 0.845 & 0.8412 & 0.665 & 0.5519 & 0.84  & 0.7919 & 0.875 & 0.849 & 0.755 & 0.5752 & 0.855 & 0.6884 & 0.875 & 0.8515 & 0.7875  & 0.7005  \\
    \hline
    FedMD & 0.692 & 0.5721 & 0.823 & 0.8027 & 0.704 & 0.6087 & 0.875 & 0.8544 & 0.907 & 0.8745 & 0.762 & 0.6627 & 0.835 & 0.6493 & 0.842 & 0.8001 & 0.8050  & 0.7281  \\
    FedDF & 0.721 & 0.5949 & 0.817 & 0.8094 & 0.723 & 0.6221 & 0.893 & 0.8735 & 0.935 & 0.9021 & 0.757 & 0.6609 & 0.847 & 0.6819 & 0.833 & 0.7826 & 0.8158  & 0.7409  \\
    pFedDF & 0.755 & 0.6536 & 0.853 & 0.8256 & 0.741 & 0.6237 & 0.894 & 0.8742 & 0.935 & 0.9021 & 0.796 & 0.7219 & 0.879 & 0.7095 & 0.874 & 0.8521 & 0.8409  & 0.7703  \\
    DS-pFL & 0.715 & 0.6099 & 0.792 & 0.7734 & 0.765 & 0.6547 & 0.899 & 0.8792 & 0.935 & 0.9021 & 0.794 & 0.7331 & 0.853 & 0.6691 & 0.851 & 0.8266 & 0.8255  & 0.7560  \\
    KT-pFL & 0.765 & 0.6733 & 0.87  & 0.8331 & 0.755 & 0.6432 & 0.885 & 0.8621 & 0.935 & 0.9021 & 0.78  & 0.6931 & 0.865 & 0.6819 & 0.905 & 0.9023 & 0.8450  & 0.7739  \\
    MH-pFLID & 0.805 & 0.6427 & 0.945 & 0.9394 & 0.82  & 0.7604 & \textbf{0.963} & \textbf{0.9457} & \textbf{0.975} & \textbf{0.9709} & \textbf{0.815} & \textbf{0.7755} & 0.895 & 0.7287 & \textbf{0.995} & 0.9583 & 0.9016  & 0.8402  \\
    \hline
    pFLMoE(Ours) & \textbf{0.875} & \textbf{0.8745} & \textbf{0.975} & \textbf{0.9749} & \textbf{0.825} & \textbf{0.7951} & 0.945 & 0.8934 & 0.965 & 0.9458 & 0.805 & 0.7428 & \textbf{0.945} & \textbf{0.8611} & \textbf{0.995} & \textbf{0.9936} & \textbf{0.9163}  & \textbf{0.8852}  \\
    \hline
    \multicolumn{19}{c}{Ocular Disease Recognition} \\
    \hline
    \multirow{2}{*}{Method} & \multicolumn{2}{c|}{ResNet} & \multicolumn{2}{c|}{shufflenetv2} & \multicolumn{2}{c|}{ResNeXt} & \multicolumn{2}{c|}{squeezeNet} & \multicolumn{2}{c|}{SENet} & \multicolumn{2}{c|}{MobileNet} & \multicolumn{2}{c|}{DenseNet} & \multicolumn{2}{c|}{VGG} & \multicolumn{2}{c}{Average} \\
\cline{2-19}          & ACC$\uparrow$   & MF1$\uparrow$   & ACC$\uparrow$   & MF1$\uparrow$   & ACC$\uparrow$   & MF1$\uparrow$   & ACC$\uparrow$   & MF1$\uparrow$   & ACC$\uparrow$   & MF1$\uparrow$   & ACC$\uparrow$   & MF1$\uparrow$   & ACC$\uparrow$   & MF1$\uparrow$   & ACC$\uparrow$   & MF1$\uparrow$   & ACC$\uparrow$   & MF1$\uparrow$ \\
    \hline
    Only Local Training & 0.6813 & 0.5607 & 0.6438 & 0.6406 & 0.5063 & 0.5019 & 0.5625 & 0.3705 & 0.8562 & 0.8532 & 0.5813 & 0.4711 & 0.5563 & 0.5061 & 0.8938 & 0.7273 & 0.6602  & 0.5789  \\
    \hline
    FedMD & 0.5375 & 0.2945 & 0.7375 & 0.7065 & 0.475 & 0.4017 & 0.5375 & 0.1748 & 0.5375 & 0.4558 & 0.6188 & 0.4245 & 0.6438 & 0.3916 & 0.8562 & 0.6114 & 0.6180  & 0.4326  \\
    FedDF & 0.6938 & 0.6413 & 0.7688 & 0.7609 & 0.5437 & 0.5397 & 0.5688 & 0.1813 & 0.6313 & 0.6288 & 0.5375 & 0.5128 & 0.5563 & 0.5312 & 0.8938 & 0.5254 & 0.6493  & 0.5402  \\
    pFedDF & 0.7312 & 0.641 & 0.7438 & 0.7324 & 0.6062 & 0.5443 & 0.5437 & 0.4536 & 0.6562 & 0.4611 & 0.5875 & 0.5095 & 0.5437 & 0.518 & 0.9062 & 0.7708 & 0.6648  & 0.5788  \\
    DS-pFL & 0.7563 & 0.6567 & 0.7625 & 0.739 & 0.575 & 0.5652 & 0.5813 & 0.3874 & 0.8625 & 0.8625 & 0.5875 & 0.5299 & 0.5875 & 0.5394 & 0.8688 & 0.6018 & 0.6977  & 0.6102  \\
    KT-pFL & 0.7625 & 0.7144 & 0.775 & 0.7566 & 0.5125 & 0.4182 & 0.5688 & 0.3877 & 0.85  & 0.8498 & 0.6062 & 0.5078 & 0.625 & 0.4726 & 0.9187 & 0.8014 & 0.7023  & 0.6136  \\
    MH-pFLID & 0.775 & 0.6899 & 0.8188 & 0.8126 & 0.635 & 0.5652 & 0.5625 & \textbf{0.4487} & 0.9125 & 0.9114 & 0.6125 & 0.5044 & 0.6188 & 0.5756 & 0.9125 & 0.8155 & 0.7310  & 0.6654  \\
    \hline
    \model{}(Ours) & \textbf{0.8052} & \textbf{0.7354} & \textbf{0.8313} & \textbf{0.8277} & \textbf{0.6562} & \textbf{0.6552} & \textbf{0.6313} & 0.4333 & \textbf{0.9625} & \textbf{0.9625} & \textbf{0.6313} & \textbf{0.5202} & \textbf{0.6500}  & \textbf{0.5833} & \textbf{0.9500}  & \textbf{0.8529} & \textbf{0.7647}  & \textbf{0.6962}  \\
    \hline
    \end{tabular}%
  \label{tab:ddc}%
  }
\end{table*}%

\begin{table}[htbp]
  \centering
  \caption{For the medical image segmentation task, we evaluate the Dice result on Polyp dataset. The larger the better. \textbf{Bold} number means the best.
  The {\color{red!70}red} boxes represent the method of using homogeneous models. Their clients use the Unet. The {\color{blue!70}blue} boxes represent the method of using heterogeneous models in each client. The four client models are set to Unet++, Unet, ResUnet, and FCN, respectively. MH-pFLID achieves the best segmentation results.}
   \resizebox{1\linewidth}{!}{
    \begin{tabular}{c|c|c|c|c|c}
    \hline
    Method & Client1 & Client2 & Client3 & Client4 & Average \\
    \hline
    \rowcolor{red!10} FedAvg & 0.5249  & 0.4205  & 0.5676  & 0.5500  & 0.5158  \\
    \rowcolor{red!10} SCAFFOLD & 0.5244  & 0.3591  & 0.5935  & 0.5713  & 0.5121  \\
   \rowcolor{red!10} FedProx & 0.5529  & 0.4674  & 0.5403  & 0.6301  & 0.5477  \\
    \hline
   \rowcolor{red!10} Ditto & 0.5720  & 0.4644  & 0.6648  & 0.6416  & 0.5857  \\
    \rowcolor{red!10} APFL  & 0.6120  & 0.5095  & 0.6333  & 0.5892  & 0.5860  \\
   \rowcolor{red!10} LG-FedAvg & 0.6053  & 0.5062  & 0.7371  & 0.5596  & 0.6021  \\
   \rowcolor{red!10} FedRep & 0.5809  & 0.3106  & 0.7088  & 0.7023  & 0.5757  \\
    \hline
   \rowcolor{red!10} FedSM & 0.6894  & 0.6278  & 0.8021  & 0.7391  & 0.7146  \\
    \rowcolor{red!10} LC-Fed & 0.6233  & 0.4982  & 0.8217  & \textbf{0.7654}  & 0.6772  \\
    \hline
   \rowcolor{red!10} \model{} (Ours)  & \textbf{0.7918}  & \textbf{0.6882}  & \textbf{0.8808}  & 0.7644  & \textbf{0.7813}  \\
    \hline
   \rowcolor{blue!10} Only Local Training & 0.7049  & 0.4906  & 0.8079  & 0.7555  & 0.6897  \\
   \rowcolor{blue!10} MH-pFLID & 0.7565  & 0.6830  & 0.8644  & 0.7644  & 0.7671  \\
    \hline
   \rowcolor{blue!10} \model{} (Ours)  & \textbf{0.7945}  & \textbf{0.6859}  & \textbf{0.8709} & \textbf{0.7710}  & \textbf{0.7806}  \\
    \hline
    \end{tabular}%
  \label{tab:seg}%
  }
\end{table}%


\textbf{Feature space transform in MoE.} 
Before the final Mixture of Experts decision, we design a feature space transform module to transform the local features into the corresponding expert's feature space. As shown in \cref{fig_3}, the local body feature is first transformed into a common space by $W_{com}$, and then separately transformed into the corresponding expert's feature space through the respective $W_{j}$. In classification tasks, $W_{com}$ and $W_{j}$ are linear layers. For the segmentation and super-resolution tasks, $W_{com}$ and $W_{j}$ are convolutional layers. 

After feature space transform, the features in the corresponding space generate the predictions through the respective experts. We utilize the Mixture of Experts (MoE) framework to effectively aggregate these predictions. To enhance the MoE's focus on key experts, inspired by \cite{liu2020att,blecher2023moeatt}, we incorporate a cross-attention mechanism to learn the weights associated with the predictions generated by each client’s experts. It is important to note that local models not only capture essential information from their respective local datasets but also tend to inherit biases that may contribute to overfitting. By utilizing the experts from all clients as candidates and employing local features as queries to extract relevant information from the collective pool of experts, we ensure that the selected information reflects the common knowledge shared across clients. We posit that this public information, derived from diverse datasets, possesses greater generalizability, while the local biases are effectively mitigated in the selection process. Consequently, we propose the adoption of a cross-attention design to filter out local biases and enhance the overall generalization capability of the model.

The MoE is illustrated in \cref{fig_3}. We involve the local body feature denoted as $I$, and concatenate all the expert predictions as $K$ and $V$.  The feature $I$ obtains the Query feature $Q$ through a linear layer $W$. The prediction of the MoE $y_{MoE}$ is represented as:
\begin{equation} 
y_{MoE} = Attention(W(I), K, V),
\end{equation} 
where $Attention$ is the attention mechanism function \cite{vaswani2023attentionneed}. More details can be found in supplementary materials.
 

\section{Experiments}
\label{sec:experiments}

\subsection{Tasks and Datasets}

We verify the effectiveness of \model{} on 5 non-IID tasks.

\textbf{A. Medical image classification (different resolution).} We use the Breast Cancer Histopathological Image Database (BreaKHis) \cite{7312934}. We perform x2↓, x4↓, and x8↓ downsampling on the high-resolution images \cite{xie2023shisrcnet}. Each resolution of medical images is treated as a client, resulting in four clients in total. The dataset for each client was randomly divided into training and testing sets at a ratio of 7:3, following previous work. For the same image with different resolutions, they will be used in either the training set or the testing set. For the model homogeneous framework, we employed ResNet$\lbrace 5 \rbrace$. For the model heterogeneous framework, we employed ResNet$\lbrace 17,11,8,5 \rbrace$.

\textbf{B. Medical image super-resolution.} We use BreaKHis dataset \cite{7312934}. We perform x2↓, x4↓, and x8↓ downsampling on the high-resolution images \cite{xie2023shisrcnetsuperresolutionclassificationnetwork}. Each downsampled resolution of medical images is treated as a client, resulting in three clients in total. We used the RCNN \cite{dong2014learning} for the model heterogeneous framework. We used SRResNet$\lbrace 6, 12, 18 \rbrace$ \cite{ledig2017photo} for the model heterogeneous framework.

\textbf{C. Medical time-series classification.} We used the Sleep-EDF dataset \cite{goldberger2000physiobank} for the time-series classification task of three clients under non-IID distribution.  For the model homogeneous framework, we employed TCN. For the model heterogeneous framework, three clients use the TCN \cite{bai2018empirical}, Transformer \cite{2021A} and RNN \cite{xie2024trls}.

\textbf{D. Medical image classification (different label distributions).} This task includes a breast cancer classification task and an ocular disease recognition task. Similar to previous work \cite{xie2024mh}, we also designed eight clients, each using a different model. They are ResNet \cite{he2015deep}, ShuffleNetV2 \cite{ma2018shufflenet}, ResNeXt \cite{xie2017aggregated}, SqueezeNet \cite{iandola2016squeezenet}, SENet \cite{hu2018squeeze}, MobileNetV2 \cite{sandler2018mobilenetv2}, DenseNet \cite{huang2017densely}, and VGG \cite{simonyan2014very}.  We apply the same non-IID label distribution method as before to the BreaKHis and ODIR-5K datasets \cite{BHATI2023106519} across 8 clients. Specifically, the data distribution varies among clients.

\textbf{E. Medical image segmentation.} Here, we focus on polyp segmentation \cite{dong2021polyp}. The dataset consists of endoscopic images collected and annotated from four centers, with each center's dataset treats as a separate client. We employed Unet \cite{ronneberger2015u} for the model homogeneous framework. For the model heterogeneous framework, each clients adopted Unet++ \cite{zhou2019unet++}, Unet \cite{ronneberger2015u}, Res-Unet \cite{diakogiannis2020resunet}, FCN \cite{long2015fully}.
, respectively.





\subsection{Results}


\textbf{Medical image classification (different resolutions).} 
In this task, we compare \model{} with the baseline framework in two different model settings. For the model homogeneous framework, all frameworks use ResNet5. For the model heterogeneous framework, we use the ResNet family. As in previous work, we use the ResNet family for the model heterogeneous framework. Clients with low-resolution images employ shallower models, while clients with high-resolution images use more complex models. In \cref{tab:drc}, experimental results show that in both model settings, \model{} achieves the best performance. This indicates that \model{} can effectively integrate knowledge from both homogeneous or heterogeneous models, thereby enhancing the performance of local models.



\textbf{Medical image super-resolution.}
This task involves reconstructing different low-resolution medical images into high-resolution images. We consider all images of the same resolution as a single client. In this task, we use the RCNN for the model homogeneous framework and the SResNet family for the model homogeneous framework. As shown in \cref{tab:SR}, \model{} achieves the best results. Moreover, as shown in \cref{fig_SR}, our framework can recover more details.

\textbf{Time-series classification.}
The experimental results in \cref{tab:tsc} show that \model{} achieves the best results under two different model settings. This further demonstrates the superiority of \model{} in federated learning of homogeneous and heterogeneous models.



\textbf{Medical image classification (different label distributions).} 
In \cref{tab:ddc}, the experimental results for the medical image classification task with different label distributions, where each client uses heterogeneous models, show that \model{} achieves the optimal results. This demonstrates that, compared to heterogeneous federated learning methods, the Mixture of Experts approach of \model{} can more effectively fuse knowledge from other clients to make decisions. 

\textbf{Medical image segmentation.} 
We validate the effectiveness of \model{} in medical image segmentation tasks. \cref{tab:seg} presents the results of federated learning in the segmentation task, demonstrating that \model{} achieves the best experimental outcomes under two different model settings. The experimental results not only demonstrate that \model{} effectively enhances local model performance, but also prove its applicability to various medical tasks. Meanwhile, the visualization results in \cref{fig_seg} show that the segmentation results of \model{} are closer to ground truth.




\begin{table}[htbp]
  \centering
  \caption{The disconnect experiment of \model{} and MH-pFLID (centralized Federated Learning) in medical image classification (different resolutions) and medical image segmentation tasks. In the communication disconnect, we randomly disconnect each client's upload or download operations with the server. At a disconnect rate of 50\%, centralized federated learning ensures that each client maintains at least one upload or download operation. At a dropout rate of 75\%, it becomes only local training. In the client disconnect, we directly remove certain clients during the federated learning process. For example, a disconnect rate of 25\% indicates that only three clients participate in the federated learning, while "None (3 clients)" refers to the performance of three clients out of four.  \model{} shows less performance degradation compared to the centralized approach in disconnect scenarios.}
  \resizebox{1\linewidth}{!}{
    \begin{tabular}{c|c|ccc|c}
    \toprule
    \multicolumn{6}{c}{Communication disconnect} \\
    \hline
    \multirow{2}{*}{Task} & \multirow{2}{*}{Method} & \multicolumn{1}{c|}{None} & \multicolumn{1}{c|}{25\%} & \multicolumn{1}{c|}{50\%} & 75\% \\
\cline{3-6}          &       & \multicolumn{1}{c|}{ACC} & \multicolumn{1}{c|}{ACC} & ACC   & ACC \\
    \hline
    \multirow{2}{*}{Classification} & \model{} (Ours)   & \multicolumn{1}{c|}{0.8880 } & \multicolumn{1}{c|}{0.8798 } & 0.8474  & 0.8011  \\
          & MH-pFLID & \multicolumn{1}{c|}{0.8583 } & \multicolumn{1}{c|}{0.8393 } & 0.7687  & 0.7258  \\
    \hline
    \multicolumn{1}{r}{} &       & \multicolumn{1}{c|}{Dice} & \multicolumn{1}{c|}{Dice} & Dice  & Dice \\
    \hline
    \multirow{2}{*}{Segmentation} & \model{} (Ours)   & \multicolumn{1}{c|}{0.7860 } & \multicolumn{1}{c|}{0.7789} & 0.7423 & 0.7211 \\
          & MH-pFLID & \multicolumn{1}{c|}{0.7671} & \multicolumn{1}{c|}{0.7641} & 0.7043 & 0.6897 \\
    \hline
    \multicolumn{6}{c}{Client disconnect} \\
    \hline
    \multirow{2}{*}{Task} & \multirow{2}{*}{Method} & \multicolumn{1}{c|}{\makecell{None \\ (3 client)}} & \multicolumn{1}{c|}{25\%} & \makecell{None \\ (2 client)} & 50\% \\
\cline{3-6}          &       & \multicolumn{1}{c|}{ACC} & \multicolumn{1}{c|}{ACC} & ACC   & ACC \\
    \hline
    \multirow{2}[1]{*}{Classification} & \model{} (Ours)   & \multicolumn{1}{c|}{0.8771 } & \multicolumn{1}{c|}{0.8633 } & \multicolumn{1}{c|}{0.8638 } & 0.8474  \\
          & MH-pFLID & \multicolumn{1}{c|}{0.8447} & \multicolumn{1}{c|}{0.8193 } & 0.8340  & 0.8087  \\
    \hline
    \multicolumn{1}{r}{} &       & \multicolumn{1}{c|}{Dice} & \multicolumn{1}{c|}{Dice} & Dice  & Dice \\
    \hline
    \multirow{2}[2]{*}{Segmentation} & \model{} (Ours)   & \multicolumn{1}{c|} {0.7873}  & \multicolumn{1}{c|}{0.7642}  & 0.7838 & 0.7446 \\
          & MH-pFLID & \multicolumn{1}{c|}{0.7681} & \multicolumn{1}{c|}{0.7359} & 0.7737 & 0.7154 \\
   \hline
    \end{tabular}%
  \label{tab:dpt}%
  }
\end{table}%

\begin{table}[htbp]

  \centering
  \Large
  \caption{Difference between communication disconnect and client disconnect incentralized and decentralized federated learning.}
  \resizebox{1\linewidth}{!}{
    \begin{tabular}{c|m{15em}|m{15em}}
    \hline
    & \makecell[c]{Centralized} & \makecell[c]{Decentralized} \\
   \hline
   \makecell{Communication \\ disconnect} & Randomly disconnect each client's upload or download operations with the central server. & Randomly disconnect the upload or download operations between each client. \\
    \hline
    \makecell{Client \\ disconnect} &  \multicolumn{2}{c} {Remove the corresponding clients during the federated learning process.}\\
    \hline
    \end{tabular}%
  \label{tab:des}%
  }
\end{table}%

\begin{table}[htbp]
  \centering
  \caption{In heterogeneous model settings, we compare the impact of expert parameter quantities on performance in medical image segmentation, time-series classification, breast cancer classification (with different label distributions), and medical super-resolution tasks. \#Params represents the average amount of parameters a client needs to share in one communication. The experimental results show that using the entire local model as experts leads to limited performance improvement.}
   \resizebox{0.9\linewidth}{!}{
    \begin{tabular}{c|cc|cc}
    \hline
    \multirow{2}{*}{Expert} & \multicolumn{2}{c|}{Segmentation} & \multicolumn{2}{c}{Time-series} \\
    \cline{2-5}       & \#Params(M) & Dice  & \#Params(M) & ACC \\
    \hline
    Head  & 0.001 & 0.7806 & 0.002 & 0.9136 \\
    Entire local model & 24.015 & 0.7921 & 1.181 & 0.9122 \\
    \hline
    \multirow{2}{*}{Expert} & \multicolumn{2}{c|}{Breast Cancer} & \multicolumn{2}{c}{Super-resolution} \\
\cline{2-5}          & \#Params(M) & ACC   & \#Params(M) & PSNR \\
    \hline
    Head  & 0.004 & 0.9163 & 0.001 & 29.30  \\
    Entire local model & 9.763 & 0.9077 & 7.321 & 29.43 \\
    \hline
    \end{tabular}%
  \label{tab:num}%
  }
\end{table}%

\begin{table}[htbp]
  \centering
  \caption{The ablation experiments of \model{}. We remove some essential modules to verify the effectiveness of each module. We perform experiments on Time-series classification, medical image super-resolution, and segmentation tasks. We observe that though those experiments outperform centralized methods, they suffer different levels of performance decrease. (MoE: Mixture of Experts; FST: Feature space transform)}
  \resizebox{1\linewidth}{!}{
    \begin{tabular}{c|cc|cc|c}
    \hline
    \multirow{2}{*}{Methods} & \multicolumn{2}{c|}{Time-series} & \multicolumn{2}{c|}{Super-resolution} & Segmentation \\
      \cline{2-6}    & ACC$\uparrow$   & MF1$\uparrow$   & PSNR$\uparrow$  & SSIM$\uparrow$  & Dice$\uparrow$ \\
    \hline
    \model{} (Ours)  &\textbf{0.9041}  & \textbf{0.8879}  & \textbf{29.30}  & \textbf{0.8350}  & \textbf{0.7860}  \\
    \hline
    w/o MoE module & 0.8731  & 0.8533  & 28.65  & 0.8234  & 0.7344  \\
    w/o  FST module & 0.8812  & 0.8681  & 28.44  & 0.8261  & 0.7421  \\
    w/ centralized MoE\& FST & 0.8609  & 0.8347  & 27.46  & 0.8199  & 0.6625  \\
    w/ aggregated head & 0.8361  & 0.8065  & 26.07  & 0.7891  & 0.5893  \\
    \hline
    \end{tabular}%
  \label{tab:ab}%
  }
\end{table}%


\textbf{Connection robustness.} As shown in \cref{tab:des}, We design two disconnect experiments for medical image classification (different resolutions) and medical image segmentation tasks to verify that \model{} can still help improve local model training performance in disconnect scenarios. Communication disconnect refers to randomly dropping clients' upload or download processes. Client disconnect means that the corresponding client does not participate in the federated learning. The experimental results are shown in \cref{tab:dpt}. In the communication disconnect experiment, the results show that compared to centralized solutions, our method experiences lower performance degradation as the dropout rate increases. When the disconnect rate reaches 75$\%$, the centralized solution performs similarly to only local training, while our approach allows for knowledge transfer, thereby enhancing local model performance. In the client disconnect experiment, \model{} still shows less performance degradation compared to the centralized approach.

\textbf{Experts number of pararmeters.}
As shown in \cref{tab:num}, under heterogeneous model settings, we compare the impact of expert parameter quantity on performance in four tasks. The experimental results show that using the entire model as an expert leads to limited performance improvements, but significantly increases the average parameter quantity that each client needs to share, resulting in a higher communication burden.

\textbf{Ablation studies.}
To verify the effectiveness of the proposed components in \model{}, a comparison between \model{} and its four components on time-series classification, super-resolution, and segmentation tasks is given in \cref{tab:ab}. The four components are as follows: (1) w/o MoE: we replace our designed MoE with the original MoE. (2) w/o FST indicates that we delete the feature space transform module in the local network. (3) w/ centralized MoE\& FST or w/ aggregated head means that all clients' MoE and FST or head parameters are uploaded to the central server for aggregation. 
Experimental results show that our designed MoE and FT modules effectively integrate knowledge from various clients. Compared to centralized aggregation, our decentralized approach better utilizes knowledge from other clients to enhance local model performance.

\section{Conclusions}
Centralized Federated Learning could lead to knowledge damage during aggregation, and the knowledge would be undermined before it reaches back to each client. It also creates a dependency on the central server, which may affect training stability if the server malfunctions or connections are unstable.
We design a decentralized federated learning framework named \model{} to address the issues of centralized Federated Learning.
\model{} shares each client's head model as an expert with other clients and uses the MoE approach to fuse the knowledge from these experts to make the final decision. We demonstrate the effectiveness of our framework in 5 Non-IID medical tasks under two model settings and achieves state-of-the-art performance.


\appendix

\section*{Appendix}

\section{Tasks and Datasets}

We verify the effectiveness of \model{} on 5 non-IID tasks. Here, we provide additional details about the tasks and the datasets used.

\textbf{A. Medical image classification (different resolution).} We use the Breast Cancer Histopathological Image Database (BreaKHis) \cite{7312934}. We treat the original image as a high-resolution image. Then, the Bicubic downsampling method is used to downsample the high-resolution image, obtaining images with resolutions of x2↓, x4↓, and x8↓, respectively. Each resolution of medical images was treated as a separate client, resulting in four clients in total. Each client has the same number of images with consistent label distribution, but the image resolution is different for each client. The dataset for each client was randomly divided into training and testing sets at a ratio of 7:3, following previous work. In this task, we employed a family of models such as ResNet$\lbrace 17,11,8,5 \rbrace$.

\textbf{B. Medical image super-resolution.} We use BreaKHis dataset \cite{7312934}. We perform x2↓, x4↓, and x8↓ Bicubic downsampling methods on the high-resolution images \cite{xie2023shisrcnetsuperresolutionclassificationnetwork}. Each downsampled resolution of medical images is treated as a client, resulting in three clients in total. The dataset for each client was randomly divided into training and testing sets at a ratio of 7:3, following previous work. We used the RCNN \cite{dong2014learning} for the model heterogeneous framework. We used SRResNet$\lbrace 6, 12, 18 \rbrace$ \cite{ledig2017photo} for the model heterogeneous framework. 

\textbf{C. Medical time-series classification.} We used the Sleep-EDF dataset 
 \cite{goldberger2000physiobank} for the classification task of time series under Non-IID distribution. We divided the Sleep-EDF dataset evenly among three clients.  The ratio of the training set to the testing set for each client is 8:2. We designed three clients using the TCN \cite{bai2018empirical}, Transformer \cite{2021A} and RNN \cite{xie2024trls}.

\textbf{D. Medical image classification (different label distributions).} This task includes a breast cancer classification task and an OCT disease classification task. We designed eight clients, each corresponding to a distinct heterogeneous model. These models included ResNet \cite{he2015deep}, ShuffleNetV2 \cite{ma2018shufflenet}, ResNeXt \cite{xie2017aggregated}, SqueezeNet \cite{iandola2016squeezenet}, SENet \cite{hu2018squeeze}, MobileNetV2 \cite{sandler2018mobilenetv2}, DenseNet \cite{huang2017densely}, and VGG \cite{simonyan2014very}. Similar to the previous approach, we applied non-IID label distribution methods to the BreaKHis (breast cancer classification) \cite{kermany2018identifying} and ODIR-5K (ocular disease recognition) across the 8 clients.

For the breast cancer classification task, we have filled in the data quantity to 8000 and allocated 1000 pieces of data to each client. The ratio of training set to testing set for each client is 8:2.

For theocular disease recognition task, we randomly selected 6400 pieces of data, with 800 pieces per client. The ratio of training set to test set is also 8:2.

\textbf{E. Medical image segmentation.} Here, we focus on polyp segmentation \cite{dong2021polyp}. The dataset for this task consisted of endoscopic images collected and annotated from four different centers, with each center's dataset treated as a separate client. Thus, there were four clients in total for this task. The number of each client are 1000, 380, 196 and 612. The ratio of the training set to the testing set for each client is 1:1.
Each client utilized a specific model, including Unet++ \cite{zhou2019unet++}, FCN \cite{long2015fully}, Unet \cite{ronneberger2015u}, and Res-Unet \cite{diakogiannis2020resunet}.

\section{Implementation Details}

For different tasks, \model{} adopts different learning rates of two-stage and batch size. The specific settings are shown in \cref{tab:cc}. In experiments, all frameworks have a communication round of 100.  For classification, $\mathcal{L}_{\textit{loc}}$ and $\mathcal{L}_{\textit{MoE}}$ are cross-entropy loss. For super-resolution tasks, $\mathcal{L}_{\textit{loc}}$ and $\mathcal{L}_{\textit{MoE}}$ are $L1$ loss. And for segmentation tasks, $\mathcal{L}_{\textit{loc}}$ and $\mathcal{L}_{\textit{MoE}}$ are Dice and cross-entropy loss. $\lambda_{\textit{loc}}$ and $\lambda_{\textit{MoE}}$ are set to 0.5. The performance evaluation of the classification task is based on two metrics, accuracy (ACC) and macro-averaged F1-score (MF1), providing a comprehensive assessment of the model's robustness. For super-resolution, we adopted the Peak-Signal to Noise Ratio (PSNR) and structural similarity index (SSIM)  to evaluate the performance. Additionally, Dice is used to evaluate the segmentation task performance across frameworks.

\begin{table*}[htbp]
  \centering
  \caption{The two-stage learning rates and batch size of \model{} under 5 tasks.}
  \resizebox{1\linewidth}{!}{
    \begin{tabular}{>{\centering\arraybackslash}m{3.4cm}|>{\centering\arraybackslash}m{4.2cm}|>{\centering\arraybackslash}m{2.6cm}|>{\centering\arraybackslash}m{3cm}|>{\centering\arraybackslash}m{4.4cm}|>{\centering\arraybackslash}m{2.3cm}}
    \hline
          & \textbf{Medical image classification (different resolution)} & \textbf{Medical image super-resolution} & \textbf{Medical time-series classification} & \textbf{Medical image classification (different label distributions)} & \textbf{Medical image segmentation}\\
    \hline
    Learning rate of  Local Network Training & 0.0005 & 0.0001 & 0.001 & 0.001 & 0.001 \\
    \hline
    Learning rate of  Mixture of Experts Decision & 0.0001 & 0.00001 & 0.0001 & 0.0001 & 0.0001 \\
    \hline
     Batch size & 32    & 16    & 256   & 8     & 8 \\
    \hline
    \end{tabular}%
  \label{tab:cc}%
  }
\end{table*}%

\section{Baselines}

In the medical image classification task (different resolution), we selected FedAvg, SCAFFOLD, FedProx,  FedRep, LG-FedAvg, APFL, and Ditto with homogeneous models. We chose MH-pFLID, FedMD, FedDF, pFedDF, DS-pFL, and KT-pFL for heterogeneous model federated learning.

For medical image super-resolution, we compared various approaches, including local training of clients and a variety of personalized federated learning techniques, as well as methods for learning a single global model. Among the personalized methods, we also chose FedRep, LG-FedAvg, APFL, and Ditto. We also compare MH-pFLID with our method (\model{}) under the heterogeneous model setting.

The baseline used in the medical time-series classification task is the same as the medical image classification task (different label distrbutions).

For image segmentation tasks, we compared various approaches, including local training of clients and a variety of personalized federated learning techniques, as well as methods for learning a single global model. Among the personalized methods, we also chose FedRep, LG-FedAvg, APFL, and Ditto. We simultaneously added LC-Fed \cite{lcfed} and FedSM \cite{fedsm} which are effective improvements for FedRep and APFL in the federated segmentation domain. We also compare MH-pFLID with our method (\model{}) under the heterogeneous model setting.

In the medical image classification task (different label distrbutions), we compared various methods, including local training of clients with heterogeneous models and existing heterogeneous model federated learning approaches (FedMD, FedDF, pFedDF, DS-pFL, and KT-pFL, MH-pFLID).

\section{Training Settings}

\subsection{Evaluation Indicators}
The performance evaluation of the classification task is based on two metrics, accuracy (ACC) and macro-averaged F1-score (MF1), providing a comprehensive assessment of the model's robustness. For the super-resolution task, we adopted the Peak Signal-to-Noise Ratio (PSNR) and structural similarity index (SSIM) to evaluate the performance. Additionally, Dice is used to evaluate the segmentation task performance across frameworks. 

\textbf{A. Accuracy.} Accuracy is the ratio of the number of correct judgments to the total number of judgments. 

\textbf{B. Macro-averaged F1-score.} First, calculate the F1-score for each recognition category, and then calculate the overall average value. 

\textbf{C. Peak Signal-to-Noise Ratio.} The formula for Peak Signal-to-Noise Ratio ($PSNR$) is typically written as:

\begin{small} 
\begin{equation} 
PSNR = 10*log_{10} (\frac {R^2}{MSE}),
\end{equation} 
\end{small}
where $R$ is the maximum possible pixel value in the image (for example, for an 8-bit image, $R$=255). $MSE$ is the Mean Squared Error, calculated as:

\begin{small} 
\begin{equation} 
MSE = \frac {1}{N}\sum_{i=1}^N(I(i)-K(i))^2, 
\end{equation} 
\end{small}
where $I(i)$ and $K(i)$ are the pixel values of the original image and the reconstructed image at position $i$, and $N$ is the total number of pixels in the image.

\textbf{D. Structural similarity index (SSIM).} First, calculate the F1-score for each recognition category, and then calculate the overall average value. 

\begin{small} 
\begin{equation} 
SSIM(x, y) = \frac{(2\mu_x\mu_y + C_1)(2\sigma_{xy} + C_2)}{(\mu_x^2 + \mu_y^2 + C_1)(\sigma_x^2 + \sigma_y^2 + C_2)},
\end{equation} 
\end{small}
where $x$ and $y$ are the two images being compared, $\mu_x$ and $\mu_y$ are the average luminance of $x$ and $y$. $\sigma_x^2$ and $\sigma_y^2$ are the variances of $x$ and $y$. $\sigma_{xy}$ is the covariance between $x$ and $y$. $C_1$ and $C_2$ are small constants to stabilize the division (typically $C_1=(K_1L)^2$ and $C_2=(K_2L)^2$, where $L$ is the dynamic range of the pixel values). 
 
\textbf{E. Dice.} It is a set similarity metric commonly used to calculate the similarity between two samples, with a threshold of [0,1]. In medical images, it is often used for image segmentation, with the best segmentation result being 1 and the worst result being 0. The Dice coefficient calculation formula is as follows:
\begin{small} 
\begin{equation} 
Dice = \frac{{2*(pred \cap true)}}{{pred \cup true}}
\end{equation} 
\end{small}
Among them, $pred$ is the set of predicted values, $true$ is the set of groudtruth values. And the numerator is the intersection between $pred$ and $true$. Multiplying by 2 is due to the repeated calculation of common elements between $pred$ and $true$ in the denominator. The denominator is the union of $pred$ and $true$.

\subsection{Loss Function}
Many loss functions have been applied in this article, and here are some explanations for them.The cross entropy loss function is very common and will not be explained in detail here. 
We mainly explain Dice loss.

Dice Loss applied in the field of image segmentation. It is represented as:

\begin{small} 
\begin{equation} 
DiceLoss = 1- \frac{{2*(pred \cap true)}}{{pred \cup true}}
\end{equation} 
\end{small}

The Dice loss and Dice coefficient are the same thing, and their relationship is:

\begin{small} 
\begin{equation} 
DiceLoss = 1- Dice
\end{equation} 
\end{small}

In super-resolution task, we use $L_1$ loss to optimize the model.

\subsection{Public Datasets for other Federated Learning of Heterogeneous Models}

In this section, we mainly describe the setting of public datasets for methods such as FedMD, FedDF, DS-pFL and KT-pFL.

\textbf{A. Medical image classification (different resolution).} We select 100 pieces of data from each client and put them into the central server as public data, totaling 400 pieces of data as public data. In order to better obtain soft predictions for individual clients, the image resolution of the publicly available dataset will be resized to the corresponding resolution for each client.

\textbf{B. Medical image classification (different label distributions).} For the breast cancer classification task, we select 50 pieces of data for each client to upload, and the public dataset contains 400 images. For the Ocular Disease Recognition task,  we also select 50 pieces of data for each client to upload, and the public dataset contains 400 images.

\textbf{C. Medical time-series classification.} We select 200 pieces of data for each client to upload, and the public dataset contains 600 images.

\section{Future Works} 
Our method has shown its effectiveness in traditional medical classification and segmentation tasks such as \cite{li2025automatic,li2024anatomask}. Moving forward, as the need for data privacy and safety concerns increases, we intend to broaden the decentralized federated learning applicability to various other computer vision tasks, such as image/video/3D understanding/generation~\cite{luan2023high,luan2021pc,luan2024spectrum}, to preserve data privacy and resolve safety concerns. 


{
    \small
    \bibliographystyle{ieeenat_fullname}
    \bibliography{main}
}

\end{document}